\documentclass[conference]{IEEEtran}
\IEEEoverridecommandlockouts

\usepackage{cite}
\usepackage{amsmath,amssymb,amsfonts}
\usepackage{algorithmic}
\usepackage{graphicx}
\usepackage{textcomp}
\usepackage{xcolor}
\usepackage{multirow}
\def\BibTeX{{\rm B\kern-.05em{\sc i\kern-.025em b}\kern-.08em
    T\kern-.1667em\lower.7ex\hbox{E}\kern-.125emX}}
\begin{document}

\title{Vision-based Fight Detection from \\ Surveillance Cameras\\
{\footnotesize 
}
\thanks{}
}

\author{\IEEEauthorblockN{\c{S}eymanur Akt{\i}}
\IEEEauthorblockA{\textit{Department of Computer Engineering} \\
\textit{ Istanbul Technical University }\\
Istanbul, Turkey \\
akti15@itu.edu.tr}
\and
\IEEEauthorblockN{G\"ozde Ay\c{s}e Tataro\u{g}lu}
\IEEEauthorblockA{\textit{Department of Computer Engineering} \\
\textit{Istanbul Technical University}\\
\textit{Idea Technology Solutions R\&D Center}\\
Istanbul, Turkey \\
tataroglu15@itu.edu.tr}
\and
\IEEEauthorblockN{Haz{\i}m Kemal Ekenel}
\IEEEauthorblockA{\textit{Department of Computer Engineering} \\
\textit{Istanbul Technical University}\\
Istanbul, Turkey \\
ekenel@itu.edu.tr}
}
\IEEEoverridecommandlockouts
\IEEEpubid{\makebox[\columnwidth]{978-1-7281-3975-3/19/\$31.00~\copyright2019 IEEE
\hfill} \hspace{\columnsep}\makebox[\columnwidth]{ }}
\maketitle
\IEEEpubidadjcol
\begin{abstract}
Vision-based action recognition is one of the most challenging research topics of computer vision and pattern recognition.
A specific application of it, namely, detecting fights from surveillance cameras in public areas, prisons, etc., 
is desired to quickly get under control these violent incidents. This paper addresses this research problem and explores LSTM-based approaches to solve it. 
Moreover, the attention layer is also utilized. 
Besides, a new dataset is collected, which consists of fight scenes from surveillance camera videos available at YouTube. This dataset is made publicly available\footnote{https://github.com/sayibet/fight-detection-surv-dataset}. From the extensive experiments conducted on Hockey Fight, Peliculas, and the newly collected fight datasets, it is observed that the proposed approach, which integrates Xception model, Bi-LSTM, and attention, improves the state-of-the-art accuracy for fight scene classification. 
\end{abstract}


\section{INTRODUCTION}
Violence detection has been receiving increasing attention as a research topic, since it has many practical use cases. 
Since, unfortunately, the violent scenes in movies or media have become common, and since young generation can have access to these media content easily, a group of research activities is on automatic detection of violent activities in media contents. 
Another main use case is to detect violent activities in public areas, such as underground, streets, buses, hospitals, welfare institutions, etc. in order to automatically warn the public officers and enable quick action against them. Violent activities contain a broad range of activities, for example, vandalism, explosion, and fighting. In this study, we focus on the fight activity. A fight event is defined as two or more people, who are fighting to a degree that must be interfered.



Related approaches consist of two parts as feature extraction and classification. Mainly two different approaches are applied for feature extraction: computing optical flow information of the videos and computing deep convolutional neural networks-based representations. Due to the proven success of convolutional neural networks (CNN) in various computer vision applications, CNN based approaches are highly preferred in recent works. 
Long Short-term Memories (LSTM) are used for modeling the temporal information, as they find out relationships between the consecutive frames through their memory ability. In summary, CNN + LSTM network is commonly used in action recognition due its high performance. 

In this study, in order to enhance the CNN + LSTM based approach for the fight detection task, a modified Xception CNN is trained using the fight scenes. Thus, it is expected that this CNN is more familiar with the input sequences and extracts more relevant features from them. In the classification layer, a novel approach is developed by using Bidirectional LSTM (Bi-LSTM) along with a self-attention layer to improve the performance. Furthermore, a new surveillance camera fight dataset is collected.

The remainder of the paper is organized as follows. Section~2 gives an overview of the related work. In section~3, technical details of the proposed method are explained. Section~4 presents and discusses the experimental results. The obtained results are summarized in section~5 and finally, the paper is concluded in section~6.

\section{RELATED WORK}
One of the most common deep learning solutions for action recognition is two-stream convolutional networks \cite{simonyan2014two}. In this method two CNNs are used, one for spatial feature extraction, which learns the actions from single images and the other one is for the temporal feature extraction, which learns from the optical flow vectors of multiple frames. Then, outputs of the two networks are combined at the end. Sudhakaran and Lanz preferred to use convolutional LSTM for classification in order to discriminate the spatio-temporal changes between frames in a better way \cite{sudhakaran2017learning}. 

Xu et al. use attention in image captioning by focusing on the objects that can give important information about what is happening in the scene \cite{xu2015show}. Sharma et al. use attention in action recognition for processing the features, which have the largest effect on the output 
\cite{sharma2015action}. In this work, GoogLeNet~\cite{googlenet} is used for feature extraction and multi-layered deep LSTM with attention mechanism is used for classification. According to experimental results, the attention layer enhances the performance of the LSTM. Song et al. apply LSTM to the skeleton data, where the subjects in video sequences are represented as skeletons to recognize the human actions. Furthermore, they benefit from the attention layer in order to focus on the most active joints of sample skeletons in terms of spatio-temporal changes between frames \cite{song2017end}. 

Liu et al. introduced a new type of LSTM, which is named as Global Context-Aware Attention LSTM \cite{liu2017global}. This new method is developed to perform 3D action recognition on skeleton data and it aims to choose the most informative joints of the samples by using an iterative attention method. 
Additionally, it evaluates the global context while learning from the frames, differently from the regular 2D LSTM. Dong et al. detected the violent actions between people by using multi-stream CNNs \cite{dong2016multi}. Firstly, CNNs extract spatio-temporal features, then they add one more stream for learning the acceleration of the videos. Thus, the sequences can be classified considering the activity of the scene \cite{dong2016multi}. Singh et al. extracted different kinds of features from video sequences through a multi-stream CNN \cite{singh2016multi}. After detecting the person in the frame, they construct a bounding box on the tracked person and use several streams for taking motion features from both inside of bounding box and general frame. Then the features are fed into a bi-directional LSTM for classifying the actions. Ullah et al. used various CNN architectures to extract features from the frames of video sequences \cite{ullah2017action}. Features are taken from the second to the last layer of network and classified by a bi-directional LSTM. 
3D convolutional neural networks are also utilized  for action recognition in video sequences \cite{ding2014violence, baccouche2011sequential, ji20123d, wang2016beyond}. Peixoto et al. used 3D CNN and CNN-LSTM for violence detection in videos. Then, they combined the outputs of these two networks with another network which can distinguish the different concepts of the violence \cite{peixoto2019}.

In the literature, there are several publicly available violence detection datasets. For example, Technicolor presents their Hollywood movie dataset that contains violent and non-violent sequences from 31 movies \cite{demarty2014benchmarking}. 
Peliculas dataset contains various fight and non-fight videos from YouTube or the movies \cite{nievas2011violence}. Hockey dataset includes fight and non-fight videos from ice hockey games \cite{nievas2011violence}. Another dataset is Violent Flows Dataset and it contains multiple violence scenes \cite{hassner2012vif}. UCF-Crimes dataset includes different crime scenarios such as robbery, argon, burglary etc. along with fighting \cite{sultani2018ucfcrime}. A recent dataset released in 2019 \cite{perez2019}, contains surveillance camera videos with fight instances. 
To complement these datasets, in this study, a fight dataset is constructed by using the surveillance camera footages from YouTube.

\section{PROPOSED METHOD}

In the following subsections, feature extraction and classification parts of the proposed method are presented.


\subsection{Feature Extraction Model}\label{AA}
Various types of CNN architectures are tested for feature extraction part, such as VGG16 \cite{vgg16} and Xception \cite{xception}.

VGG16 takes 224 x 224 pixel resolution images as input. 
It has three fully connected layers at the end. The features are taken from the second fully connected layer. On the other hand, Xception takes 299 x 299 pixel resolution input. 
The features are extracted from the last global average pooling layer.

Furthermore, one additional CNN is trained for fight detection, which is named as Fight-CNN. Fight and non-fight frames of the video sequences in Hockey dataset are used for training. The trained CNN has the Xception architecture but the last layer is mapped into two classes. Also the kernel size is widened in order to catch more relative features from the fight scenes. The new network with Xception is smaller than the regular model with 11 million parameters. It has two fully connected layers before classification layer and features are extracted from the first fully connected layer.

Before sending the videos for the feature extraction, frames are sampled from video sequences. 
Uniform sampling is used and 5 or 10 frames from each video are selected. 
Then, using cubic interpolation these frames are resized to the input size of the network architecture. 

\subsection{Classification Model}\label{AA}
In the classification part, Bi-LSTM is used, since it can learn the dependency between past and current information. Then, an attention layer is included to determine the significant parts of the input.

1) LSTM: Long Short-term Memory is a method that is used in sequence learning tasks \cite{hochreiter1997long}. 
The memory usage capability of LSTM differs from the regular recurrent neural networks (RNN). Its memory gates in the modules make it possible to keep the necessary information and ignore irrelevant information. The gates choose to pass or throw some parts of the data according to its relevancy by considering the previous data. In other words, the gates in LSTM learn how much the new information depends on the previous information. Therefore, the relationship between the elements of a sequence can be learned. In this case, the data consists of sequence of images and the network can connect the information in frames which are taken at different times from the videos. During this process, the system remembers the previous frame while examining the current frame. The system learns the temporal changes occurring during the video processing and those changes give significant information to recognize the actions.


During the LSTM experiments, an LSTM model with one LSTM layer, three dense (1024, 50, 2) and three activation layers (relu, sigmoid, softmax) are used. At the end of the architecture, softmax layer is used with two classes instead of binary classification by sigmoid. Therefore, the prediction confidences in the output can be observed. So that, mean squared error is used as the loss function which gives better results than the cross entropy loss function.

\begin{figure*}[!htbp]
\centering
     \includegraphics[width=0.98\textwidth]{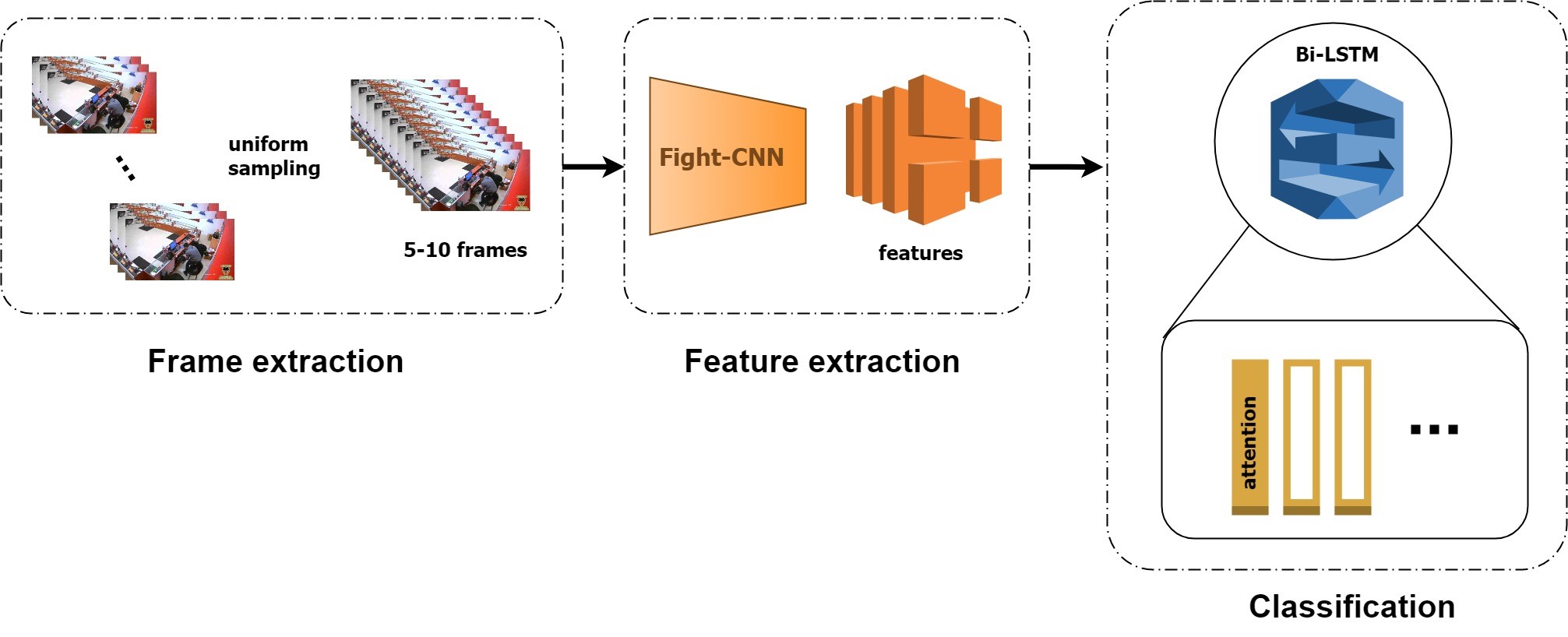}
      \caption{Overview diagram of the proposed system.}
       \label{fig}
\end{figure*}

2) Bi-LSTM: Different from the regular LSTM which has only forward flow in the sequence where the inputs are determined according to the previous information, Bi-LSTM has an additional backward flow \cite{schuster1997bidirectional}. After completing the forward learning, a backward learning is processed starting from the last element to the first element. Therefore, in each cell, both the past and future information is kept and outputs are determined by taking into account this information.

While performing the experiments with Bi-LSTM, the same architecture with regular LSTM is used with an additional Bi-LSTM layer instead of LSTM layer. Besides, dropout is applied in order to reduce overfitting.

3) Attention layer: Attention mechanism is first introduced by Bahdanau et al. in 2014 \cite{bahdanau2014neural} and generally used in natural language processing in RNNs for deciding how much attention must be given to other words while processing the current word. It is also used in visual problems like image captioning \cite{anderson2018bottom, you2016image, lu2017knowing} and object detection \cite{ba2015multiple}.

When attention layer is used together with bi-directional LSTMs, it computes weights for each cell to interpret each element in the sequence. The backward and forward layer values of each element is calculated and affect the other elements’ outputs. Attention layer determines how much each output should be affected by other inputs. After observing both past and future information, it generates a weight matrix and this matrix is used to calculate the outputs.

Self-attention \cite{vaswani2017attention} is another type of the attention mechanism, which is used in this study. The authors apply the attention to the input data and try to represent it in a more convenient form by focusing on significant parts of the data while processing the elements in sequence. For instance, the input data in this study is feature vectors from ten frames. The attention layer performs on the input and generates new feature vectors considering the attention matrix and relationships between input vectors. After that, the new feature vectors are sent into the next layers for classification. The overview of the proposed system can be seen in Fig. \ref{fig}.

\section{EXPERIMENTAL RESULTS}

In the following subsections, we first explain the used datasets and the experimental setups. Then, we present and discuss the experimental results.

\subsection{Datasets}\label{AA}

\begin{table}[]
\caption{Number of samples for each dataset}
\resizebox{0.48\textwidth}{!}{
\begin{tabular}{|l|c|c|c|}
\hline
\multicolumn{1}{|c|}{Datasets}        & \# fight & \# non-fight & \# total \\ \hline
Hockey Dataset                        & 500      & 500          & 1000     \\ \hline
Peliculas Dataset                     & 100      & 100          & 200      \\ \hline
Collected Surveillance Camera Dataset & 150      & 150          & 300      \\ \hline
\end{tabular}
}
\end{table}

1) Hockey Fight Dataset: The dataset contains fight and non-fight scenes from ice hockey games. There are 1000 video samples in total, where 500 of them are fight sequences and other 500 of them are non-fight sequences. Videos are two seconds long and frame sizes are constant. Background of the videos are all similar and they contain background motion.

 2) Peliculas Dataset: It includes fight sequences from Hollywood movies, some non-fight scenes from football games, and other events. There are 200 videos in total. 100 of them are fight videos and 100 of them are non-fight videos. Duration of videos are two seconds and size of the frames can differ. Environment and people in the videos are varying, since they are from the movie scenes. These videos also have background motion.

\begin{figure}[h!]
\centerline{\includegraphics[width=0.5\textwidth]{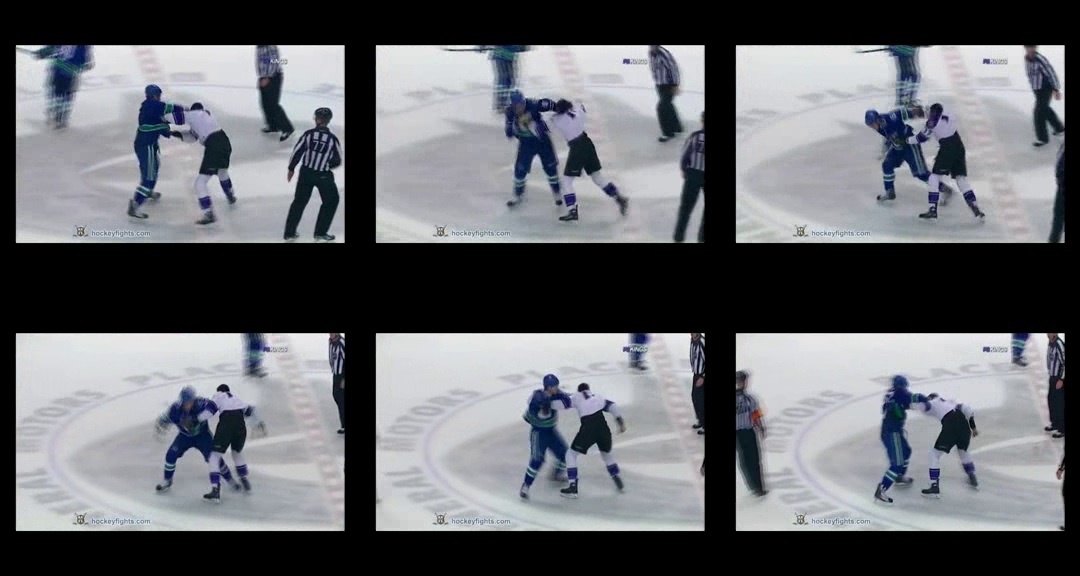}}
\caption{An example fight scene from Hockey dataset.}
\label{hockey-image}
\end{figure}

\begin{figure}[h!]
\centerline{\includegraphics[width=0.5\textwidth]{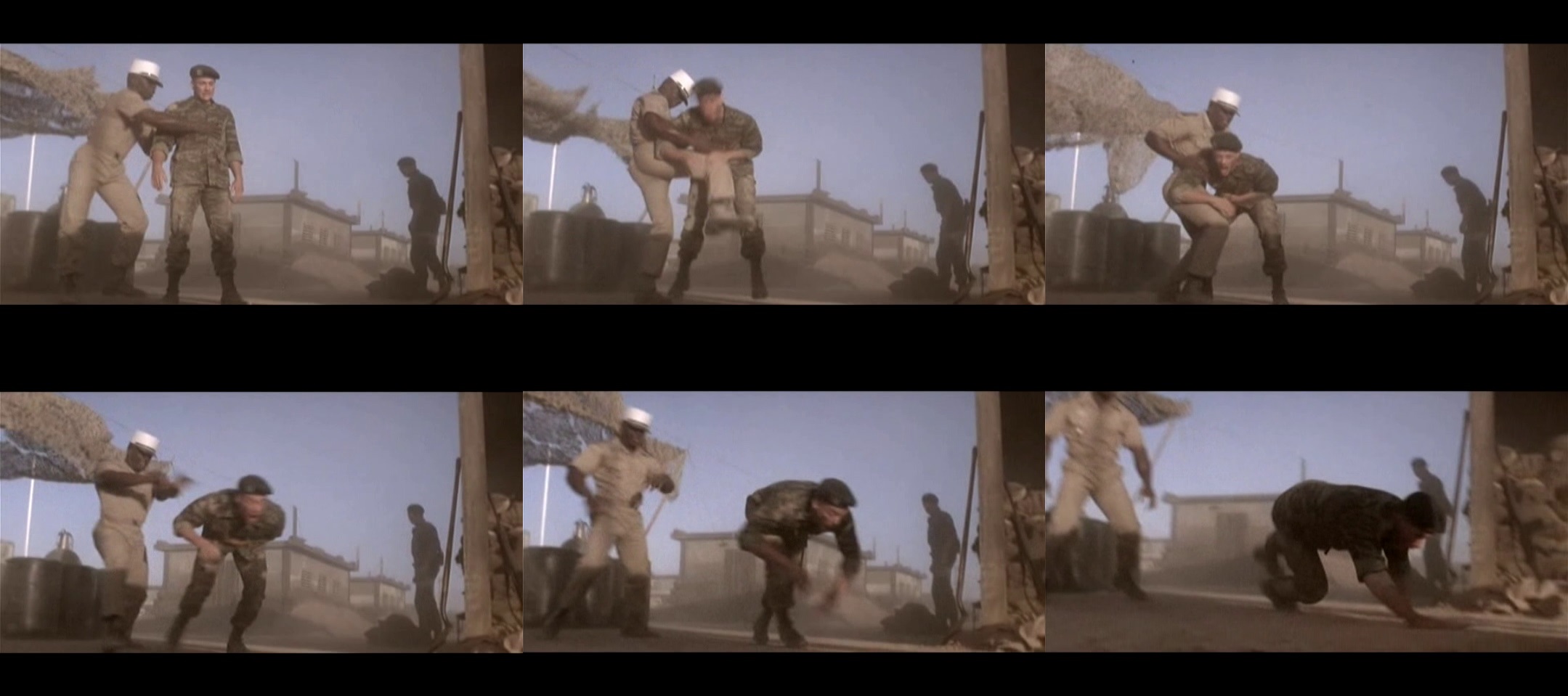}}
\caption{An example fight scene from Peliculas dataset.}
\label{pel-image}
\end{figure}

\begin{figure}[h!]
\centerline{\includegraphics[width=0.5\textwidth]{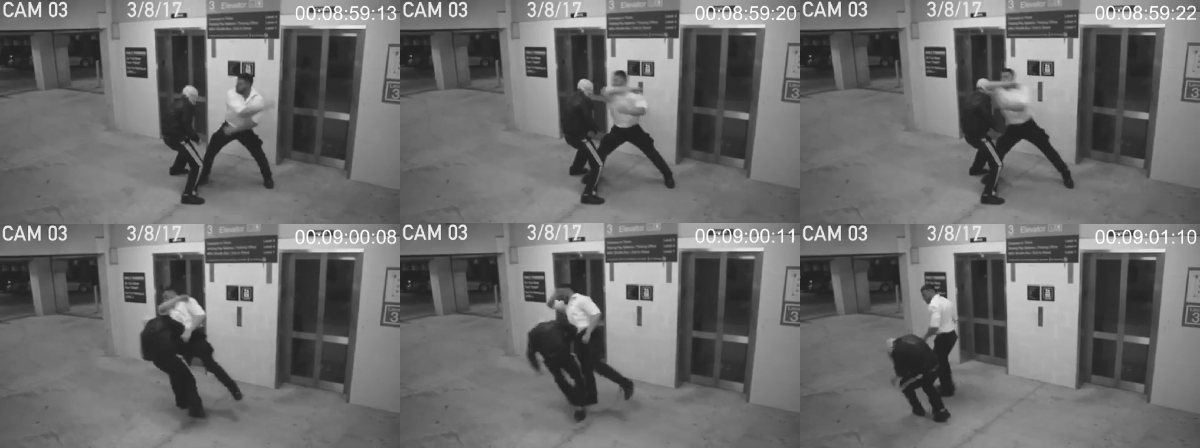}}
\caption{An example fight scene from surveillance camera dataset.}
\label{fig3}
\end{figure}

3) Surveillance Camera Fight Dataset: This dataset is collected for this study. Even though there are some fight or violence specific datasets, the main samples in these datasets are taken from movies or hockey games, which correspond to different type of scenes. These datasets can help to learn actions itself, but they are not exactly suitable for the purposed task. The actors in the hockey game scene records look identical and the background itself does not change much. However, in surveillance applications, humans in the scenes always differ and the background of the footage differs for each camera. In movies and hockey games, the background is moving due to filming techniques like zoom in / out. On the other hand, surveillance cameras are mostly still and the background in recordings is more stable. The differences can be observed from Fig. \ref{hockey-image}, \ref{pel-image}, \ref{fig3}. Thus, a new dataset containing the fight / non-fight sequences from surveillance camera footage would complement the existing datasets. 

In surveillance camera dataset, there are 300 videos in total, 150 of them are fight sequences and 150 of them are non-fight sequences. The surveillance camera footages are collected from YouTube mostly and some surveillance camera datasets like CamNet \cite{zhang2015camera} and Synopsis dataset \cite{wang2017event, huang2014maximuma} are used for extracting non-fight video cuts. After collecting videos, 2-second-long fight / non-fight sequences are cut from them. The videos have different sizes and different number of frames. Therefore, the frames are resized before they are sent to the CNNs. Then uniform sampling is applied by taking into account the total frame number of the videos as seen in Fig. \ref{fig}. 
Table 1 summarizes the number of samples in the used datasets.

\begin{figure}[h!]
\centerline{\includegraphics[width=0.45\textwidth]{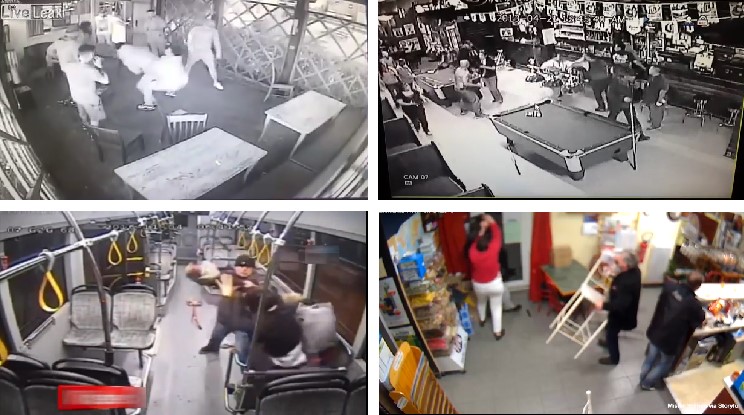}}
\caption{Various fight scenarios from the collected dataset.}
\label{fig4}
\end{figure}

There are various types of fight scenarios in the dataset such as kick, fist, hitting with an object, and wrestling. 
Since the security camera footages contain different light and coloring conditions, these variations are also taken into consideration to increase the diversity in the dataset further. In addition, security camera footages from different places are collected like cafe, bar, street, bus, shops, etc. This way, the variety in the dataset is ensured. Fight scenarios are independent from the environment of the surveillance camera as seen in Fig. \ref{fig4}.

This dataset is publicly available and can be accessed through https://github.com/sayibet/fight-detection-surv-dataset.

\subsection{Results}
Each experiment is conducted for each three datasets: Hockey, Peliculas, and surveillance camera dataset. For feature extraction part, VGG16 and Xception architectures are tested. In addition, a modified Xception architecture is trained using the fight scenes from Hockey dataset and named as Fight-CNN.

For the classification part, regular LSTMs and Bi-LSTMs are tested along with VGG16 and Xception models. Also the network is augmented by attention layer, which are tested by Xception and Fight-CNN. For each CNN, two classifiers which are Bi-LSTM with attention or Bi-LSTM without attention, are considered. In CNN and LSTM experiments, to observe the effect of number of frames to the accuracy, frame numbers are changed between 5 and 10. 

Number of epochs is 20, batch size is 10 for Fight-CNN experiments and 100 for VGG16 and Xception experiments. Datasets are split as 80\% for training and 20\% for testing. Experimental results are presented in terms of test accuracy in Tables 2-3-4.

\begin{table}[]
\centering
\caption{Experimental results on Peliculas dataset.}
\label{tab:my-table}
\begin{tabular}{|c|l|c|c|}
\hline
\multicolumn{2}{|c|}{\multirow{3}{*}{\textbf{}}}                                                & \multicolumn{2}{c|}{\textbf{Peliculas Dataset}} \\ \cline{3-4} 
\multicolumn{2}{|c|}{}                                                                          & 10 Frames              & 5 Frames               \\ \cline{3-4} 
\multicolumn{2}{|c|}{}                                                                          & \textit{accuracy}      & \textit{accuracy}      \\ \hline
\multicolumn{2}{|c|}{VGG16 + LSTM}                                                              & 95\%                   & 100\%                  \\ \hline
\multicolumn{2}{|c|}{VGG16 + Bi-LSTM}                                                           & 100\%                  & 100\%                  \\ \hline
\multicolumn{2}{|c|}{Xception + LSTM}                                                           & 97.5\%                 & 97.5\%                 \\ \hline
\multicolumn{2}{|c|}{Xception + Bi-LSTM}                                                        & 97.5\%                 & 97.5\%                 \\ \hline
\multicolumn{2}{|c|}{\textbf{Xception + Bi-LSTM + attention}}                                            & \textbf{100\%   }                  & \textbf{100\%}                  \\ \hline
\multicolumn{2}{|c|}{Fight-CNN + Bi-LSTM}                                                       & 77.5\%                     & 80\%                   \\ \hline
\multicolumn{2}{|c|}{\begin{tabular}[c]{@{}c@{}}Fight-CNN + Bi-LSTM + attention\end{tabular}} & 87.5\%                      & 90\%                   \\ \hline
\end{tabular}
\end{table}

Since Fight-CNN is trained with the scenes from Hockey dataset, the test result of the Fight-CNN on Peliculas is not as good as can be seen in Table 2. The Peliculas dataset has little amount of fight scenes samples, so the accuracy is highly affected by the false predictions. Therefore the standard deviation of accuracy is higher than others. At the end of the training, loss values of Bi-LSTM methods are mostly lower than the regular LSTM models. As it is observed in Table~2, addition of the attention layer significantly increases the accuracy compared to the other approaches.

\begin{table}[]
\centering
\caption{Experimental results on Hockey dataset.}
\label{tab:my-table}
\begin{tabular}{|c|l|c|c|}
\hline
\multicolumn{2}{|c|}{\multirow{3}{*}{\textbf{}}}                                                & \multicolumn{2}{c|}{\textbf{Hockey Dataset}} \\ \cline{3-4} 
\multicolumn{2}{|c|}{}                                                                          & 10 Frames              & 5 Frames               \\ \cline{3-4} 
\multicolumn{2}{|c|}{}                                                                          & \textit{accuracy}      & \textit{accuracy}      \\ \hline
\multicolumn{2}{|c|}{VGG16 + LSTM}                                                              & 87.05\%                   & 92.5\%                  \\ \hline
\multicolumn{2}{|c|}{VGG16 + Bi-LSTM}                                                           & 92.5\%                  & 91\%                  \\ \hline
\multicolumn{2}{|c|}{Xception + LSTM}                                                           & 93.5\%                 & 93.5\%                 \\ \hline
\multicolumn{2}{|c|}{Xception + Bi-LSTM}                                                        & 94.5\%                 & 95\%                 \\ \hline
\multicolumn{2}{|c|}{\textbf{Xception + Bi-LSTM + attention}}                                            & \textbf{97.5\%}                     & \textbf{98\% }                \\ \hline 
\multicolumn{2}{|c|}{Fight-CNN + Bi-LSTM}                                                       & 95.5                     & 93.5\%                   \\ \hline
\multicolumn{2}{|c|}{\begin{tabular}[c]{@{}c@{}}Fight-CNN + Bi-LSTM + attention\end{tabular}} & 96\%                     & 95\%                   \\ \hline
\end{tabular}
\end{table}

The Hockey dataset experiments indicate the advantage of Bi-LSTMs over regular LSTMs as seen in Table 3. The attention layer shows its effect again when it is compared with the Xception and Fight-CNN experiments. The results of Fight-CNN along with Bi-LSTM and attention are found to be promising. 
Since the Xception network that we use in Fight-CNN structured with few parameters, it gives lower accuracy compared with regular Xception network. On the other hand, Fight-CNN contains less number of parameters and extracts features faster than regular Xception network.


%
\begin{table}[]
\centering
\caption{Experimental results on collected surveillance camera dataset.}
\label{tab:my-table}
\resizebox{0.48\textwidth}{!}{
\begin{tabular}{|c|l|c|c|}
\hline
\multicolumn{2}{|c|}{\multirow{3}{*}{\textbf{}}}                                                & \multicolumn{2}{c|}{\textbf{Surveillance Camera Fight Dataset}} \\ \cline{3-4} 
\multicolumn{2}{|c|}{}                                                                          & 10 Frames              & 5 Frames               \\ \cline{3-4} 
\multicolumn{2}{|c|}{}                                                                          & \textit{accuracy}      & \textit{accuracy}      \\ \hline
\multicolumn{2}{|c|}{VGG16 + LSTM}                                                              & 62\%                   & 61.67\%                  \\ \hline
\multicolumn{2}{|c|}{VGG16 + Bi-LSTM}                                                           & 45\%                  & 52\%                  \\ \hline
\multicolumn{2}{|c|}{Xception + LSTM}                                                           & 60\%                 & 55\%                 \\ \hline
\multicolumn{2}{|c|}{Xception + Bi-LSTM}                                                        & 63.3\%                 & 63\%                 \\ \hline
\multicolumn{2}{|c|}{Xception + Bi-LSTM + attention}                                            & 69\%                     & 68\%                  \\ \hline
\multicolumn{2}{|c|}{Fight-CNN + Bi-LSTM}                                                       & 68.5                     & 70\%                   \\ \hline
\multicolumn{2}{|c|}{\begin{tabular}[c]{@{}c@{}}\textbf{Fight-CNN + Bi-LSTM + attention}\end{tabular}} & \textbf{71\% }                    &\textbf{ 72\%}                   \\ \hline
\end{tabular}
}
\end{table}

As can be seen in Table 4, the results for surveillance camera dataset is not as good as the ones presented for the other datasets. Since the variety of the samples in this dataset is very high, the models cannot easily generalize to this dataset. The results show that Fight-CNN provides a better feature extraction on the data, when it is compared to Xception model. 
Since the CNN is familiar with the fight scenes that it is trained with, it can extract the significant features more easily. Again the attention layer increased the accuracy in both regular Xception and Fight-CNN with its focusing ability. 


It is observed that the number of frames per video parameter has no direct correlation with the accuracy in most of the cases. However, using five frames per video has less computation load for the feature extraction step compared with using ten frames per video.

\section{DISCUSSION}
The proposed method has benefited from the CNNs for feature extraction from frames. Two-way learning of bi-directional LSTMs and the attention layers that can also determine the amount of given attention to each part of the sequence are found to improve the accuracy. As a result, proposed method has surpassed the state-of-the-art performance. 
Additionally, a new model is tested by using Fight-CNN, a modified version of Xception model. 

Bi-LSTMs show better performance than regular LSTMs in action recognition, as also stated in related studies in \cite{singh2016multi, dong2016multi}. Also the studies in \cite{xu2015show, sharma2015action, song2017end} show that the attention layer improves the performance of sequence learning. This study validates this finding and shows that using Bi-LSTM together with attention is a promising solution to classify fight scenes. 

The experimental results also indicate that the more diversity a dataset contains, the more challenging it gets to classify fight scenes. Since the collected surveillance fight dataset contains different types of fight events, from different locations, under different conditions, it poses a significant challenge for the state-of-the-art action recognition systems.

\section{CONCLUSION}
The main objective of this study is detecting fight scenes from surveillance cameras in a fast and accurate way. 
The proposed method which employs attention layer along with Bi-LSTM networks has improved the detection accuracy and provided promising results. Moreover, using a pre-trained Fight-CNN for feature extraction proves its effectiveness on surveillance camera dataset experiments. 

Another important contribution of the study is the collected surveillance camera fight dataset, which presents further challenges for automatic fight detection. This surveillance camera dataset can be extended by adding new samples from security camera footages on streets or underground stations. 


\vspace{12pt}

\bibliography{ref.bib}
\bibliographystyle{IEEEtran}
\end{document}